\documentclass{article}
\usepackage{ijcai24}

\usepackage{times}
\usepackage{soul}
\usepackage{url}
\usepackage[pagebackref,breaklinks,colorlinks]{hyperref}
\usepackage[utf8]{inputenc}
\usepackage[small]{caption}
\usepackage{graphicx}
\usepackage{amsmath}
\usepackage{amsthm}
\usepackage{booktabs}
\usepackage{algorithm}
\usepackage{algorithmic}
\usepackage[switch]{lineno}
\usepackage{amssymb}
\usepackage{subfig}
\usepackage{color}
\usepackage[numbers]{natbib}

\title{StableIdentity: Inserting Anybody into Anywhere at First Sight}

\author{Qinghe~Wang$^{1}$, Xu~Jia$^{1*}$, Xiaomin~Li$^{1}$, Taiqing~Li$^{1}$, Liqian~Ma$^{2}$, Yunzhi~Zhuge$^{1}$, Huchuan~Lu$^{1}$\\
$^1$Dalian University of Technology, $^2$ZMO AI Inc.\\~\\
\url{https://qinghew.github.io/StableIdentity}
}

\begin{document}
\maketitle

\newcommand\blfootnote[1]{%
\begingroup
\renewcommand\thefootnote{}\footnote{#1}%
\addtocounter{footnote}{-1}%
\endgroup
}

\blfootnote{$^*$ Corresponding authors.}

\begin{abstract}
Recent advances in large pretrained text-to-image models have shown unprecedented capabilities for high-quality human-centric generation, however, customizing face identity is still an intractable problem. Existing methods cannot ensure stable identity preservation and flexible editability, even with several images for each subject during training. In this work, we propose StableIdentity, which allows identity-consistent recontextualization with just one face image. More specifically, we employ a face encoder with an identity prior to encode the input face, and then land the face representation into a space with an editable prior, which is constructed from celeb names. By incorporating identity prior and editability prior, the learned identity can be injected anywhere with various contexts. In addition, we design a masked two-phase diffusion loss to boost the pixel-level perception of the input face and maintain the diversity of generation. Extensive experiments demonstrate our method outperforms previous customization methods. In addition, the learned identity can be flexibly combined with the off-the-shelf modules such as ControlNet. Notably, to the best knowledge, we are the first to directly inject the identity learned from a single image into video/3D generation without finetuning. We believe that the proposed StableIdentity is an important step to unify image, video, and 3D customized generation models.
\end{abstract}

\section{Introduction}
With the boom in diffusion models~\cite{rombach2022high,ramesh2022hierarchical,zhang2023text}, customized generation has garnered widespread attention~\cite{ruiz2023hyperdreambooth,yuan2023inserting,li2023photomaker}. This task aims to inject new subject~(e.g., identity) into the text-to-image models and generate images with consistent subjects in various contexts while aligning the input text prompt. For example, users can upload their photos to obtain interesting pictures, such as ``wearing a Superman outfit". The success of customized generation can facilitate many applications such as personalized portrait photos~\cite{liu2023facechain}, virtual try-on~\cite{chen2023anydoor} and art \& design~\cite{ploennigs2023ai}.

However, existing customization methods solve this task by either finetuning the part/all parameters of the model or learning a generic encoder. Parameter finetuning methods~\cite{gal2022image,dong2022dreamartist,ruiz2023dreambooth} take a long time to search optimal parameters, but often return an inaccurate trivial solution for representing the identity. Especially if only with a single image, these methods tend to overfit the input, resulting in editability degradation. Alternatively, the encoder-based methods~\cite{ye2023ip,yan2023facestudio} require large-scale datasets for training and struggle to capture distinctive identity and details. Moreover, the identity learned by current methods is susceptible to be inconsistent with the target identity in various contexts. Therefore, there is an urgent need to propose a new framework to address the enormous challenges~(e.g., unstable identity preservation, poor editability, inefficiency) faced by this task.

Here we are particularly interested in customized generation for human under one-shot training setting, and how to store identity information into word embeddings, which can naturally integrate with text prompts. We believe prior knowledge can help for this task. On the one hand, face recognition task~\cite{wang2021deep} has been fully explored and the identity-aware ability of pretrained models can be exploited. On the other hand, text-to-image models, which are trained on massive internet data, can generate images with celeb names in various contexts, thus these names contain rich editability prior. Using these priors can alleviate these challenges, and some methods~\cite{chen2023dreamidentity,yuan2023inserting} have made preliminary attempts.

\begin{figure*}[!t]
  \centering
  \includegraphics[width=1\linewidth]{"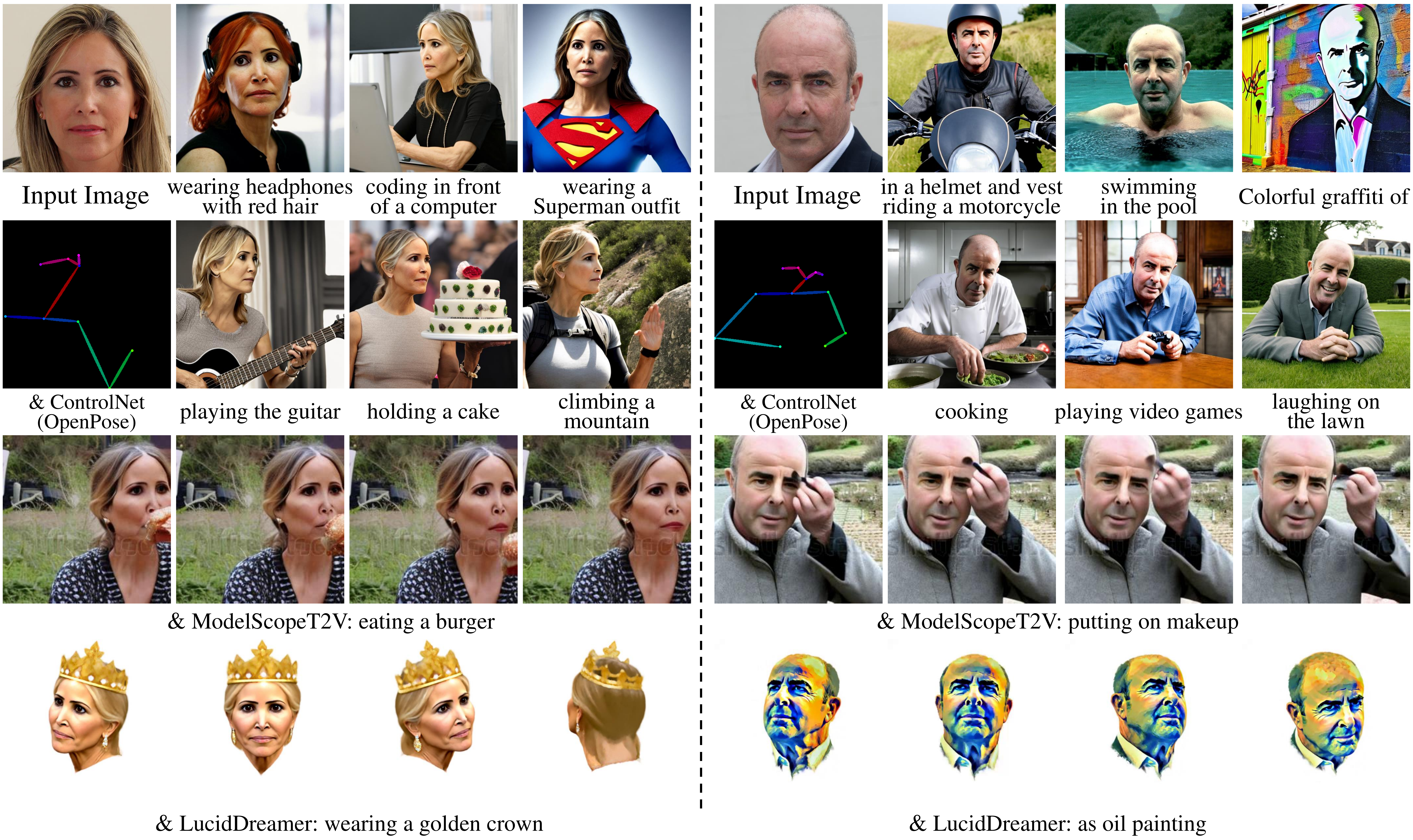"}
 \caption{Given a single input image, the proposed \textit{StableIdentity} can generate diverse customized images in various contexts. Notably, we present that the learned identity can be combined with ControlNet~\cite{zhang2023adding} and even injected into video~(ModelScopeT2V~\cite{wang2023modelscope}) and 3D~(LucidDreamer~\cite{liang2023luciddreamer}) generation.}
  \label{fig:first_image}
  \vspace{-4pt}
\end{figure*}

In this work, we propose StableIdentity which incorporates identity prior and editability prior into the human-centric customized generation. Specifically, an encoder pretrained on face recognition task is introduced to capture identity representation. Celeb names are collected to construct an embedding space as a prior identity distribution for customized generation. To encourage the target identity to perform like celeb names in pretrained diffusion model, we further land the identity representation into the prior space. Furthermore, to learn more stable identity and fine-grained reconstruction, we design a masked two-phase diffusion loss, which assigns specialized objectives in the early and late phases of denoising process respectively. Extensive experiments show StableIdentity performs favorably against state-of-the-art methods and we further analyse our superiority over several baselines of the same kind. The proposed method also shows stable generalization ability, which can directly collaborate with the off-the-shelf image/video/3D models as shown in Figure~\ref{fig:first_image}.

Our contributions can be summarized as follows:
\begin{itemize}
\item We propose StableIdentity, which incorporates identity prior and editability prior to enable identity-consistent recontextualization with just one face image.

\item We design a masked two-phase diffusion loss to perceive pixel-level details and learn more stable identity for diverse generation.

\item Extensive experiments show that our method is effective and prominent. Remarkably, our method can not only combine with image-level modules, but also unlock the generalization ability that the identity learned from a single image can achieve identity-consistent customized video/3D generation without finetuning.

\end{itemize}

\section{Related Work}
\label{related_work}
\subsection{Text-to-Image Diffusion Models}
Diffusion models~\cite{ho2020denoising,song2020denoising} have exhibited overwhelming success in text-conditioned image generation, deriving numerous classical works~\cite{rombach2022high,nichol2021glide,hertz2022prompt}. Among them, Stable Diffusion~\cite{rombach2022high} is widely used for its excellent open-source environment. In practice, Stable Diffusion can generate diverse and exquisite images from Gaussian noises and text prompts with DDIM sampling~\cite{song2020denoising}. Since the training dataset contains lots of celeb photos and corresponding names, Stable Diffusion can combine celeb names with different text prompts to generate diverse images. However, ordinary people cannot enjoy this ``privilege" directly. Therefore, to democratize Stable Diffusion to broader users, many studies~\cite{chen2023dreamidentity,yuan2023inserting,chen2023photoverse} have focused on the customized generation task.

\begin{figure*}[!t]
  \centering
  \includegraphics[width=1\linewidth]{"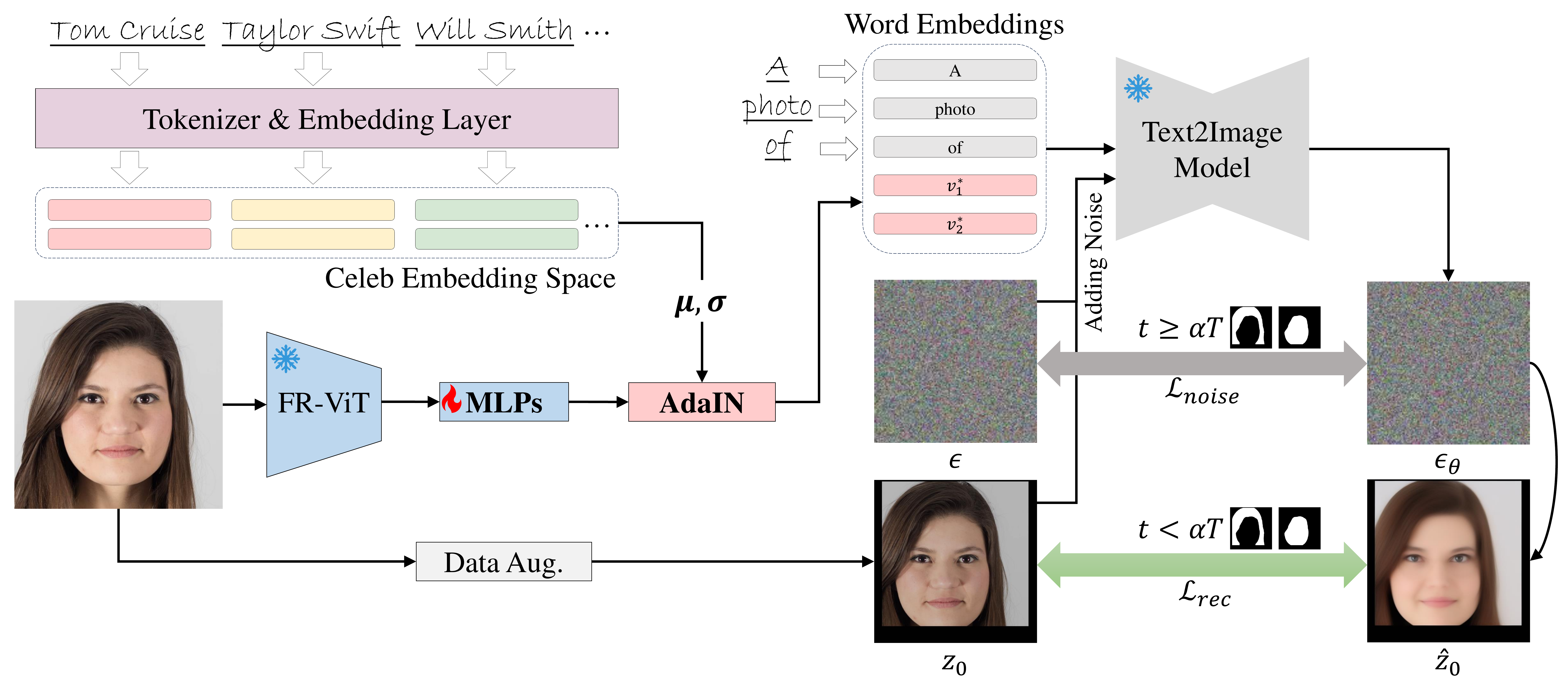"}
  \vspace{-16pt}
      \caption{Overview of the proposed \textit{StableIdentity}. Given a single face image, we first employ a FR-ViT encoder and MLPs to capture identity representation, and then land it into our constructed celeb embedding space to better learn identity-consistent editability. In addition, we design a masked two-phase diffusion loss including $\mathcal{L}_{noise}$ and $\mathcal{L}_{rec}$ for training.}
  \label{fig:framework}
\end{figure*}

\subsection{Customized Generation}
\label{sec:related_customized}
Currently, customized generation methods can be mainly divided into optimization-based and encoder-based methods. The former often require long time to optimize, while the latter need large-scale data and struggle to learn a distinctive identity. Given 3-5 images of the same subject, Textual Inversion~\cite{gal2022image} optimizes a new word embedding to represent the target subject. DreamBooth~\cite{ruiz2023dreambooth} finetunes the entire model to fit the target subject only. On the other hand, ELITE~\cite{wei2023elite}, InstantBooth~\cite{shi2023instantbooth} and IP-Adapter~\cite{ye2023ip} introduce identity information into attention layers by learning an encoder. FastComposer~\cite{xiao2023fastcomposer} trains its encoder with the whole U-Net of Stable Diffsuion together to capture identities. There are also some methods that incorporate an encoder to assist the optimization-based methods~\cite{wu2023singleinsert}, raising the performance ceiling. Celeb-Basis~\cite{yuan2023inserting} collects 691 celeb names which are editable in Stable Diffusion to build a celeb basis by PCA~\cite{pearson1901liii}. The weight of basis is optimized based on the output of ArcFace encoder~\cite{deng2019arcface}, a new identity's representation can be obtained by weighting the basis. However, the mentioned methods still perform imbalance on identity preservation and editability.

In comparison, our method exploits identity and editability prior to significantly ease the optimization process, and learns more stable identity with the proposed loss. Since Stable Diffusion is fixed, plug-and-play modules such as ControlNet~\cite{zhang2023adding} can be employed seamlessly. Furthermore, to the best knowledge, we are the first work to enable the learn identity from a single image injected into video~\cite{wang2023modelscope}~/~3D generation~\cite{liang2023luciddreamer}.

\section{Method}
Given a single face image, we aim to represent its identity via word embeddings as shown in Figure~\ref{fig:framework}, to implement identity-consistent recontextualization under various text prompts. To achieve this, we incorporate identity prior and editability prior~(See Sec.~\ref{sec:prior}) and propose a masked two-phase diffusion loss~(See Sec.~\ref{sec:masked_two_phase}).

\subsection{Preliminary}
In this work, we adopt the pretrained Stable Diffusion~\cite{rombach2022high} as our text-to-image model (denoted as SD). SD consists of three components: a VAE~($\mathcal{E}$, $\mathcal{D}$)~\cite{esser2021taming}, a denoising U-Net $\epsilon_\theta$ and a CLIP text encoder $e_{text}$~\cite{radford2021learning}. Benefiting from the high-quality reconstruction of VAE, the diffusion process of input image $x$ is performed in the latent space $z$~($z=\mathcal{E}(x)$). Specifically, at random timestep $t$~($t\in[1,1000)$), $z_t$ can be sampled as a weighted combination $z_0$ and a random noise $\epsilon$~($\epsilon\sim\mathcal{N}(0,\mathbf{I})$):
\begin{equation}
\label{add_noise}
z_t = \sqrt{\bar{\alpha}_{t}}z_0+\sqrt{1-\bar{\alpha}_t}\epsilon,
\end{equation}
where $\bar{\alpha}_{t}$ is a predefined hyperparameter set. Meanwhile, given text prompts $p$, the tokenizer of $e_{text}$ divides and encodes $p$ into $l$ integer tokens sequentially. Then, the embedding layer in $e_{text}$ obtains a corresponding embedding group $g = [v_1, ..., v_l], v_i\in\mathbb{R}^d$ which consists of $l$ word embeddings by looking up the dictionary. After that, the text transformer $\tau_{text}$ of $e_{text}$ further represents $g$ to guide model to generate images conforming to the given text prompts $p$. With latent $z_t$, the training process is optimized by:
\begin{equation}
\label{L_noise}
     \mathcal{L}_{noise} = \mathbb{E}_{z,g,\epsilon,t}\left[\|\epsilon-\epsilon_\theta(z_t,t,\tau_{text}(g))\|^2_2\right]
\end{equation}

\subsection{Model Architecture}
\label{sec:prior}
\noindent\textbf{Identity Prior.}
Existing methods extract subject information commonly with CLIP image encoder, which is pretrained for learning high-level semantics, lacks detailed identity perception. Therefore, we employ a ViT~\cite{dosovitskiy2020image} encoder finetuned for face recognition task~(denote as FR-ViT) to reap ID-aware representation $I$ from the input image.

To maintain the generalizability and editability of the vanilla SD, we fix the FR-ViT encoder and SD. Following~\cite{chen2023dreamidentity,yuan2023inserting}, we only project $I$ into two word embeddings $[v'_1, v'_2]$ with MLPs: 
\begin{equation}
\label{mlp}
     [v'_1, v'_2] = MLPs(I)
\end{equation}

Benefiting the identity prior knowledge, we can inject facial features from the input image into diffusion model more efficiently without additional feature injection.

\noindent\textbf{Editability Prior.}
Since SD is trained on large-scale internet data, using celeb names can generate images with prompt-consistent identity. Therefore, we posit that the celeb names constitute a space with editability prior. We consider 691 celeb names~\cite{yuan2023inserting} as sampling points in this space and intend to represent this space distribution with the mean and standard deviation of their word embeddings. However, in practice, the tokenizer decomposes unfamiliar word into multiple tokens~(e.g., \textit{Deschanel}~$\rightarrow[561,31328,832]$), consequently the number of tokens produced by different celeb names may not be equal. To find an editable space with a uniform dimension, we select celeb names consisting only of first name and last name, and each word corresponds to only one token~(e.g., \textit{Tom Cruise}~$\rightarrow[2435,6764]$). Eventually we obtain 326 celeb names and encode them into the corresponding word embeddings $C\in\mathbb{R}^{326\times d}$.

To master the identity-consistent recontextualization ability like celeb embeddings, we employ AdaIN~\cite{dumoulin2016learned} to incorporate the editablity prior and land $[v'_1, v'_2]$ into celeb embedding space:
\begin{equation}
\label{adain}
v^*_i = \sigma(C)(\frac{v'_i - \mu(v'_i)}{\sigma(v'_i)}) + \mu(C), for~i = 1,2
\end{equation}
where $\mu(v'_i), \sigma(v'_i)$ are scalars. $\mu(C)\in\mathbb{R}^d$, $\sigma(C)\in\mathbb{R}^d$ are vectors, since each dimension of $C$ has a different distribution. With this editablity prior, the learned embeddings $[v^*_1, v^*_2]$ are closer to the celeb embedding space than baselines as shown in Figure~\ref{fig:tsne}, which improves editablity elegantly and effectively. In addition, it also constrains the optimization process within the celeb embedding space and prevents drifting towards other categories.

\begin{figure}[!t]
  \centering
  \includegraphics[width=1\linewidth]{"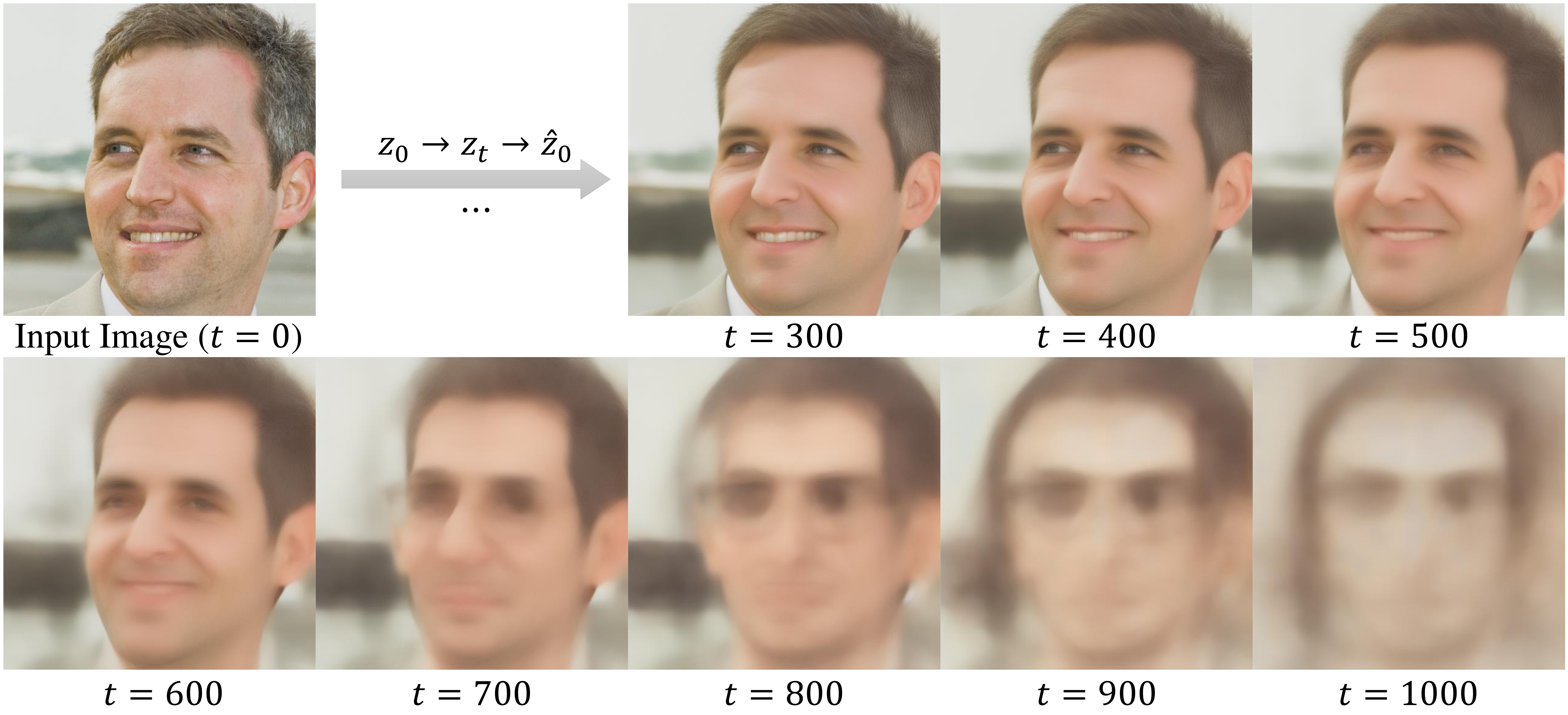"}
      \caption{We present the predicted $\hat{z}_0$ from $z_t$ at various timestep $t$. $\hat{z}_0$ at $t=\{100,200\}$, similar to $t=300$, are omitted for brevity.}
  \label{fig:pred_z0}
\end{figure}

\subsection{Model Training}
\label{sec:masked_two_phase}
\noindent\textbf{Two-Phase Diffusion Loss.}
In addition to the architecture design, we rethink the training objective of diffusion models. The vanilla training loss~$\mathcal{L}_{noise}$ excites the denoising U-Net $\epsilon_\theta$ to predict the noise $\epsilon$ contained in the input $z_t$ at any time $t$, and the introduced $\epsilon$ is randomly sampled each time. Therefore, such an objective function only implicitly and inefficiently learns the identity in the input image.

DDIM~\cite{song2020denoising} proposes a denoised observation predicted by a variant of Eq.~\ref{add_noise}: $\hat{z}_0 = \frac{z_t - \sqrt{1-\bar{\alpha}_t}\epsilon_\theta}{\sqrt{\bar{\alpha}_{t}}}$. A naive idea is to replace $\mathcal{L}_{noise}$ with the mean squared error between the predicted $\hat{z}_0$ and the real $z_0$~\cite{wu2023singleinsert}: $\mathcal{L}_{rec} = \mathbb{E}_{z,g,\epsilon,t}\left[\|\hat{z}_0-z_0\|^2_2\right]$, which can explicitly optimize the reconstruction for $z_0$. However, we observe that as timestep increases, predicted $\hat{z}_0$ becomes more difficult to approximate the true distribution of $z_0$ as shown in Figure~\ref{fig:pred_z0}. Therefore, for larger timestep, $\mathcal{L}_{rec}$ becomes less meaningful and even misleads the model to focus excessively on pixel-level reconstruction. To this end, we propose two-phase diffusion loss divided by timestep $\alpha T$:
\begin{equation}
\resizebox{.91\linewidth}{!}{$
            \displaystyle
\mathcal{L}_{diffusion} = \begin{cases}
\mathbb{E}_{z,g,\epsilon,t}\left[\|\epsilon-\epsilon_\theta(z_t,t,\tau_{text}(g))\|^2_2\right] &t\geq\alpha T, \\
\mathbb{E}_{z,g,\epsilon,t}\left[\|\hat{z}_0-z_0\|^2_2\right] &t<\alpha T.
\end{cases}
$}
\end{equation}

Empirically, the division parameter $\alpha\in[0.4,0.6]$ yields good results that balance identity preservation and diversity~($\alpha=0.6$ as default). Using $\mathcal{L}_{noise}$ at the early phase of denoising process that decides the layout of generated image~\cite{liu2023cones,xiao2023fastcomposer,mou2023t2i} can allow the learned identity to adapt to diverse layouts, while using $\mathcal{L}_{rec}$ at the late phase can boost the pixel-level perception for input image to learn more stable identity.

\noindent\textbf{Masked Diffusion Loss.}
To prevent learning irrelevant background, we also employ the masked diffusion loss~\cite{avrahami2023break,wu2023singleinsert}. Specifically, we use a pretrained face parsing model~\cite{yu2018bisenet} to obtain the face mask $M_f$ and hair mask $M_h$ of the input image. The training loss is calculated in the face area and hair area respectively:
\begin{equation}
\mathcal{L} = M_f\odot\mathcal{L}_{diffusion} + \beta M_h\odot\mathcal{L}_{diffusion}.
\end{equation}
In our experiments, we set $\beta = 0.1$ as default.

\begin{figure*}[!t]
  \centering
  \includegraphics[width=1\linewidth]{"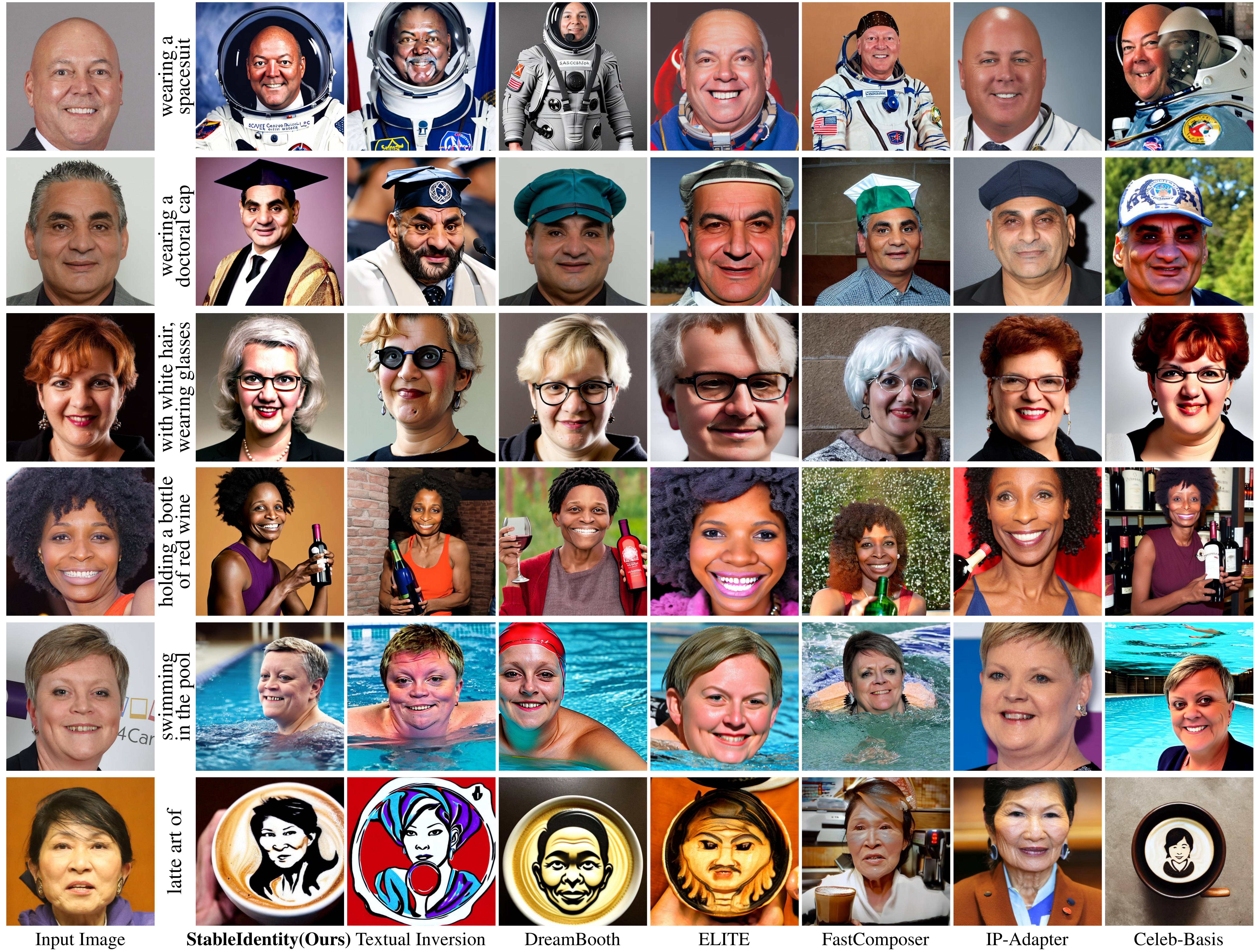"}
      \caption{We present the qualitative comparisons with six baselines for different identities~(including various races) and diverse text prompts~(covering decoration, action, attribute, background, style). Our method achieves high-quality generation with consistent identity and outstanding editability~(\textit{Zoom-in for the best view}). We provide more results in supplementary material.}
  \label{fig:comparison}
\end{figure*}

\section{Experiments}
\subsection{Experimental Setting}
\label{Exp_Set}
\noindent\textbf{Implementation Details.}
\label{sec:implement_details}
Our experiments are based on Stable Diffusion 2.1-base. The FR-ViT is a ViT-B/16 encoder finetuned for face recognition task. For an input single image, we use color jitter, random shift, random crop and random resize as data augmentations. The learning rate and batch size are set to $5e-5$ and $1$. The MLPs are trained for 450 steps~(4~mins). The placeholders $v_1^*$ $v_2^*$ of prompts such as ``$v_1^*$ $v_2^*$ wearing a spacesuit", ``latte art of $v_1^*$ $v_2^*$" are omitted for brevity in this paper. The scale of classifier-free guidance~\cite{ho2022classifier} is set to 8.5 by default. Our experiments are conducted on a single A800 GPU.

\noindent\textbf{Dataset.} We randomly select 70 non-celeb images from the FFHQ~\cite{karras2019style} and resize to $512\times512$ as our test dataset. To perform a comprehensive evaluation, we employ 40 test prompts which cover actions, decorations, attributes, expressions and backgrounds~\cite{li2023photomaker}.

\noindent\textbf{Baselines.} We compare the proposed method with baselines including the optimization-based methods: Textual Inversion~\cite{gal2022image}, DreamBooth~\cite{ruiz2023dreambooth}, Celeb-Basis~\cite{yuan2023inserting} and the encoder-based methods: ELITE~\cite{wei2023elite}, FastComposer~\cite{xiao2023fastcomposer}, IP-Adapter~\cite{ye2023ip}. We prioritize using the official model released by each method. For Textual Inversion and DreamBooth, we use their Stable Diffusion 2.1 versions for a fair comparison.

\noindent\textbf{Evaluation Metrics.} Following DreamBooth~\cite{ruiz2023dreambooth}, we calculate CLIP~\cite{radford2021learning} visual similarity~(\textbf{CLIP-I}) to evaluate high-level semantic alignment and text-image similarity~(\textbf{CLIP-T}) to measure editablity. Besides, we calculate the \textbf{Face Similarity} by ArcFace~\cite{deng2019arcface} and \textbf{Face Diversity}~\cite{li2023photomaker,wu2023singleinsert} by LPIPS~\cite{zhang2018perceptual} on detected face regions between the generated images and real images of the same ID. However, some baselines may generate completely inconsistent faces under various text prompts, which incorrectly raise face diversity. Therefore, we propose the \textbf{Trusted Face Diversity} by the product of cosine distances from face similarity and face diversity for each pair images, to evaluate whether the generated faces are both diverse and similar. To measure the quality of generation, we randomly sample 70 celeb names to generate images with test prompts as pseudo ground truths and calculate Fréchet Inception Distance (\textbf{FID})~\cite{lucic2017gans} between the generated images by the competing methods and pseudo ground truths.

\begin{table*}[!t]
\centering
\caption{Quantitative comparisons with baselines. $\uparrow$ indicates higher is better, while $\downarrow$ indicates that lower is better. The best result is shown in \underline{\textbf{bold}}. Our method obtains the best results over the text consistency (i.e., CLIP-T), identity preservation~(i.e., Face Similarity), diversity of generated faces (i.e., Trusted Face Diversity), and generation quality (i.e., FID).
 }
\label{tab:comparsion}
\setlength{\tabcolsep}{5pt}
\begin{tabular}{lcccccc}
\toprule

                          & CLIP-I$\uparrow$(\%)          & CLIP-T$\uparrow$(\%)         & Face Sim.$\uparrow$(\%) & Face Div.$\uparrow$(\%)  & Trusted Div.$\uparrow$(\%) & FID$\downarrow$                \\
\midrule                  
Textual Inversion                   & 61.30          & 28.23          & 31.30           & 37.03          & 10.75             & 28.64          \\
DreamBooth                           & 67.01          & 28.91          & 35.80           & 36.79          & 5.89             & 48.55          \\
ELITE                                & 73.94          & 26.43          & 12.58           & 25.55          & 5.35             & 84.32          \\
FastComposer & 72.32          & 28.87          & 36.91           & 28.84          & 13.90             & 47.98          \\
IP-Adapter                           & \underline{\textbf{85.14}} & 23.67          & 21.73           & 25.61          & 11.06             & 78.95          \\
Celeb-Basis                          & 63.69          & 27.84          & 25.55           & \underline{\textbf{37.85}} & 13.41             & 33.72          \\
StableIdentity~(Ours)           & 65.91          & \underline{\textbf{29.03}} & \underline{\textbf{37.12}}  & 35.46          & \underline{\textbf{15.01}}    & \underline{\textbf{24.92}}  \\
\bottomrule   
\end{tabular}
\end{table*}

\subsection{Comparison}
\noindent\textbf{Qualitative Comparison.}
As shown in Figure~\ref{fig:comparison}, given a single image as input, we show the generation results with various prompts. Textual Inversion is optimized only with $\mathcal{L}_{noise}$, which leads to a trivial solution for identity in different contexts. DreamBooth finetunes the whole SD model to fit the input face, but still fails to learn similar identities~(row~$1_{th},5_{th}$) and tends to replicate the foreground face~(row~$2_{th},3_{th}$). The encoder-based methods ELITE and IP-Adapter only learn rough shape and attributes of the input face, perform mediocrely in both identity preservation and editability. FastComposer finetunes a CLIP image encoder and the whole SD for learning identities, but suffers from low quality and artifacts~(row~$4_{th},5_{th},6_{th}$). Celeb-Basis also fails to learn accurate identity for recontextualization~(row~$1_{th},3_{th}$). Notably, when using ``latte art of" as text prompt, all baselines either produce inconsistent identity or fail to get the desired style in row~$6_{th}$. In comparison, benefiting from the effectiveness of the proposed method, our results shows superiority both in terms of identity preservation and editablity.

\noindent\textbf{Quantitative Comparison.}
In addition, we also report the quantitative comparison in Table~\ref{tab:comparsion}. Some baselines like ELITE and IP-Adapter learn only facial structure and attributes, and are susceptible to generate frontal view, resulting in better CLIP-I. This metric focuses on high-level semantic alignment and ignores identity consistency. Therefore, these methods obtain worse face similarity~(-24.54, -15.39 than ours) and trusted face diversity~(-9.66, -3.95 than ours). We also observe that the optimization-based methods Textual Inversion and DreamBooth fail to learn stable identities for recontextualization and tend to overfit to the input face, leading to poor trusted face diversity~(-4.26, -9.12 than ours). Our method achieves best performance on vision-language alignment~(CLIP-T), identity preservation~(Face Sim.), identity-consistent diversity~(Trusted Div.) and image quality~(FID).

\begin{table}[t]
\centering
\caption{Ablation study. We also present results with various division parameter $\alpha$ in the supplementary material.}
\label{tab:ablation}
\setlength{\tabcolsep}{3pt}
\begin{tabular}{lcccc}
\toprule
                 & CLIP-T$\uparrow$         & Face Sim.$\uparrow$      & Trusted Div.$\uparrow$   & FID$\downarrow$           \\
\midrule  
CLIP Enc.     & 28.03          & 35.73          & 14.81 & 25.66 \\
w/o AdaIN        & 24.81          & \underline{\textbf{47.81}}          & 13.73          & 48.73          \\
w/o Mask         & 28.15          & 34.98          & 14.47          & 25.12          \\
Only $\mathcal{L}_{noise}$    & 28.81 & 36.55 & 14.97          & 25.76          \\
Only $\mathcal{L}_{rec}$ & 27.35 & 30.69          & 13.89          & 40.54          \\
Ours             & \underline{\textbf{29.03}}          & 37.12          & \underline{\textbf{15.01}}          & \underline{\textbf{24.92}}\\
\bottomrule  
\end{tabular}
\end{table}

\subsection{Ablation Study}
\begin{figure}[!t]
  \centering
  \includegraphics[width=1\linewidth]{"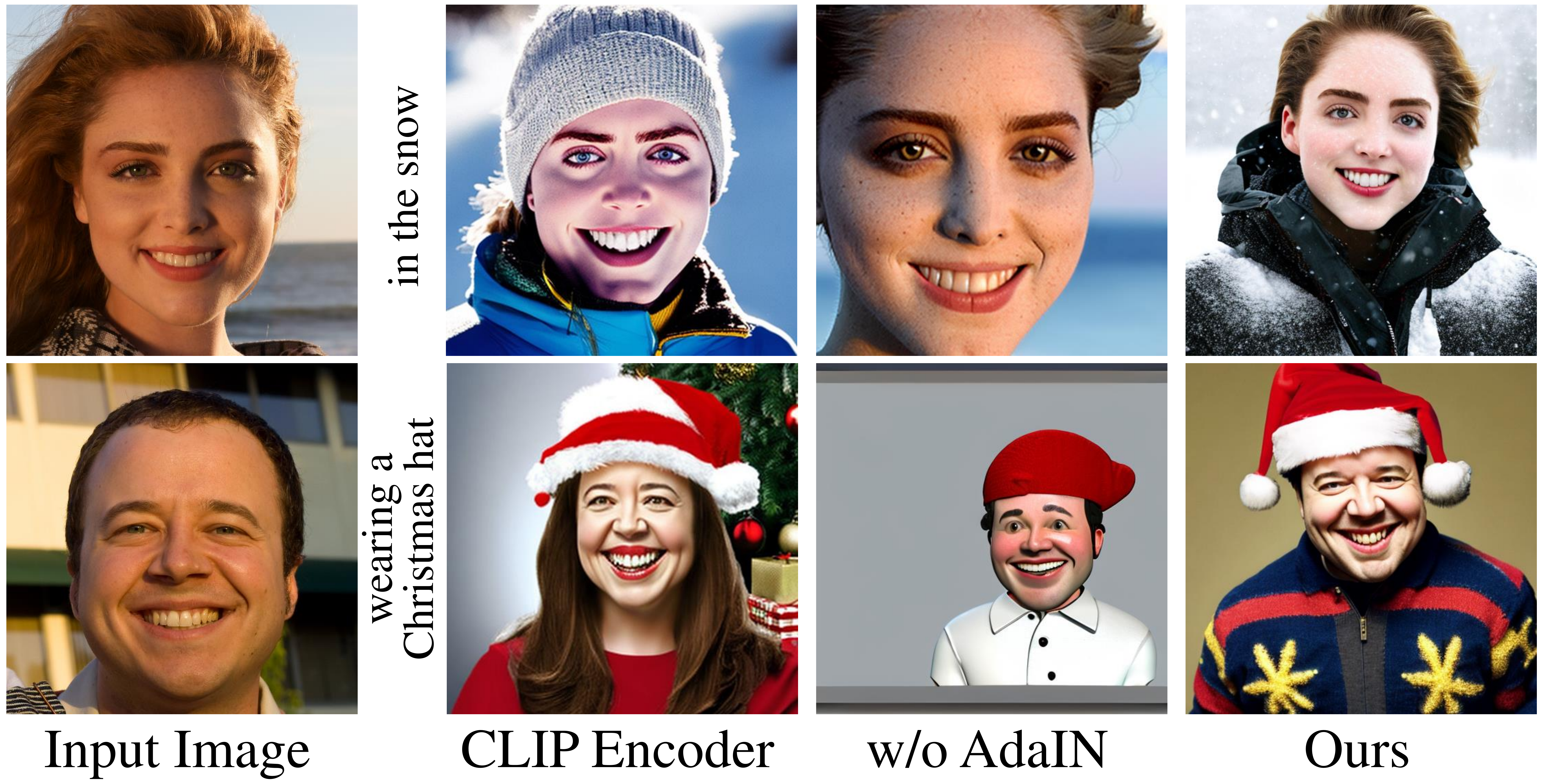"}
      \caption{Ablation study for model architecture. We show the results of using the CLIP image encoder and removing the AdaIN.}
  \label{fig:ablation_a}
\end{figure}

\begin{figure}[!t]
  \centering
  \includegraphics[width=1\linewidth]{"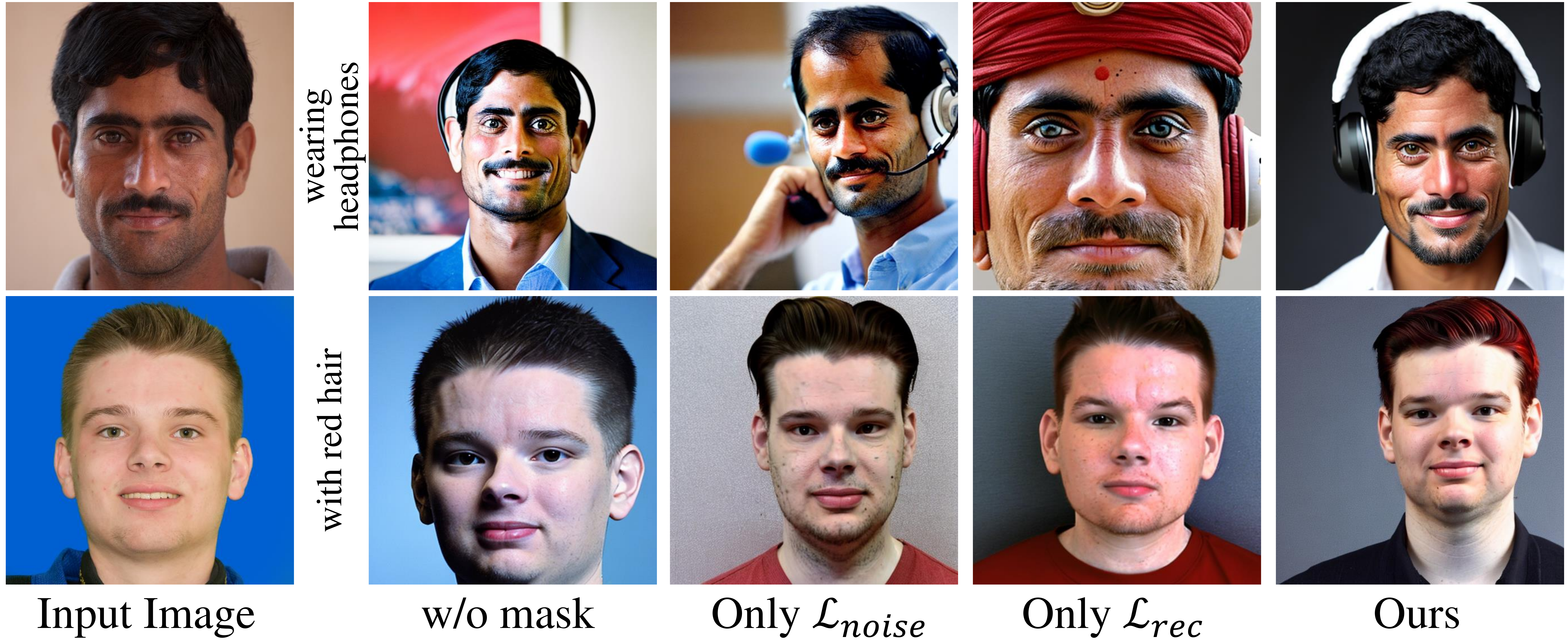"}
      \caption{Ablation study for training loss. We present the visualization results of various loss settings.}
  \label{fig:ablation_b}
\end{figure}

We conduct a comprehensive ablation study across various settings as shown in Table~\ref{tab:ablation} and Figure~\ref{fig:ablation_a},~\ref{fig:ablation_b}. We employ the CLIP Image Encoder as a baseline which is commonly adopted in encoder-based methods. Following~\cite{shi2023instantbooth,xiao2023fastcomposer}, we use the CLS feature of CLIP encoder's last layer to extract identity information. In col~2 of Figure~\ref{fig:ablation_a}, it can be observed that the CLIP image encoder is mediocre for identity preservation~(-1.39 than ours on Face Sim.). On the other hand, the setting of ``w/o AdaIN" cannot explicitly learn editability and fails to limit the value range of the learned word embeddings. It tends to generate the frontal faces and fails to align the desired text prompt~(col~3 in Figure~\ref{fig:ablation_a}), therefore it obtains high face similarity but poor CLIP-T, Trusted Div., and FID~(-4.22, -1.28, -23.81 than ours).

Furthermore, we show the ablation results for the training loss. The masked diffusion loss has been proven effective~\cite{avrahami2023break,wu2023singleinsert} and it does help focus foreground and prevent background leakage. The reconstruction of the ``Only $\mathcal{L}_{noise}$" setting is inferior than ours and is prone to undesired changes and artifacts~(col~3 in Figure~\ref{fig:ablation_b}), resulting lower identity preservation and image quality~(i.e., -0.60, -0.84 than ours on Face Sim., FID). Due to the meaningless $\mathcal{L}_{rec}$ in the early phase of denoise process, the ``Only $\mathcal{L}_{rec}$" setting only learns mediocre identities with artifacts~(col~4 in Figure~\ref{fig:ablation_b}) and leads to unsatisfactory face similarity, trusted diversity, and FID~(-6.43, -1.12, -15.62 than ours). In comparison, the proposed masked two-phase diffusion loss shows best results, and the discussion of the division parameter $\alpha$ can be found in supplementary material.

\section{Discussion}
\subsection{Downstream Applications}
\noindent\textbf{Pose-controlled Customized Image Generation.}
Since the pretrained Stable Diffusion is fixed, SD-based plug-and-play modules can collaborate with our method. ControlNet controls the pretrained SD to support additional input conditions such as keypoints, edge maps, etc. In this paper, we obtain pose images with human skeletons as condition by OpenPose~\cite{cao2017realtime}, as an example. As shown in the row~2 of Figure~\ref{fig:first_image}, we demonstrate the integration of StableIdentity and ControlNet~(SD2.1 version) which achieves simultaneous structure-controlled and identity-driven generation.

\noindent\textbf{Zero-shot Identity-driven Video/3D Generation.} Our method can be considered as introducing new identity for the dictionary of CLIP text encoder. Therefore, we believe that ideally, as long as the SD-based video and 3D generation models do not finetune the CLIP text encoder, the learned identity can be directly injected into these models.

ModelScopeT2V~\cite{wang2023modelscope} is a text-to-video generation model which brings some temporal structures into the U-Net of SD2.1 and finetunes the U-Net on large-scale datasets~\cite{schuhmann2021laion,bain2021frozen,xu2016msr}. We attempt to insert the learned identity into the unchanged CLIP text encoder without finetuning as shown in the row~3 of Figure~\ref{fig:first_image}. The generated video shows promising identity preservation and text alignment.

LucidDreamer~\cite{liang2023luciddreamer} is a text-to-3D generation pipeline based on 3D Gaussian Splatting~\cite{kerbl20233d} and allows to sample directly with the pre-trained SD2.1, like us. Therefore, it can naturally collaborate with our method. In a similar way, we insert the learned identity into this pipeline, as shown in the row~4 of Figure~\ref{fig:first_image}. The generated results achieve stable identity, high fidelity and geometry consistency. The result of ``wearing a golden crown" exhibits precise geometric structures and realistic colors and the ``as oil painting" obtains the desired style, a 3D portrait oil painting that does not exist in reality.

Overall, our method can effortlessly enable prompt-consistent identity-driven video/3D generation with the off-the-shelf text-to-video/text-to-3D models. We show more results of video/3D in the supplementary material.

\begin{figure}[!t]
  \centering
  \includegraphics[width=0.93\linewidth]{"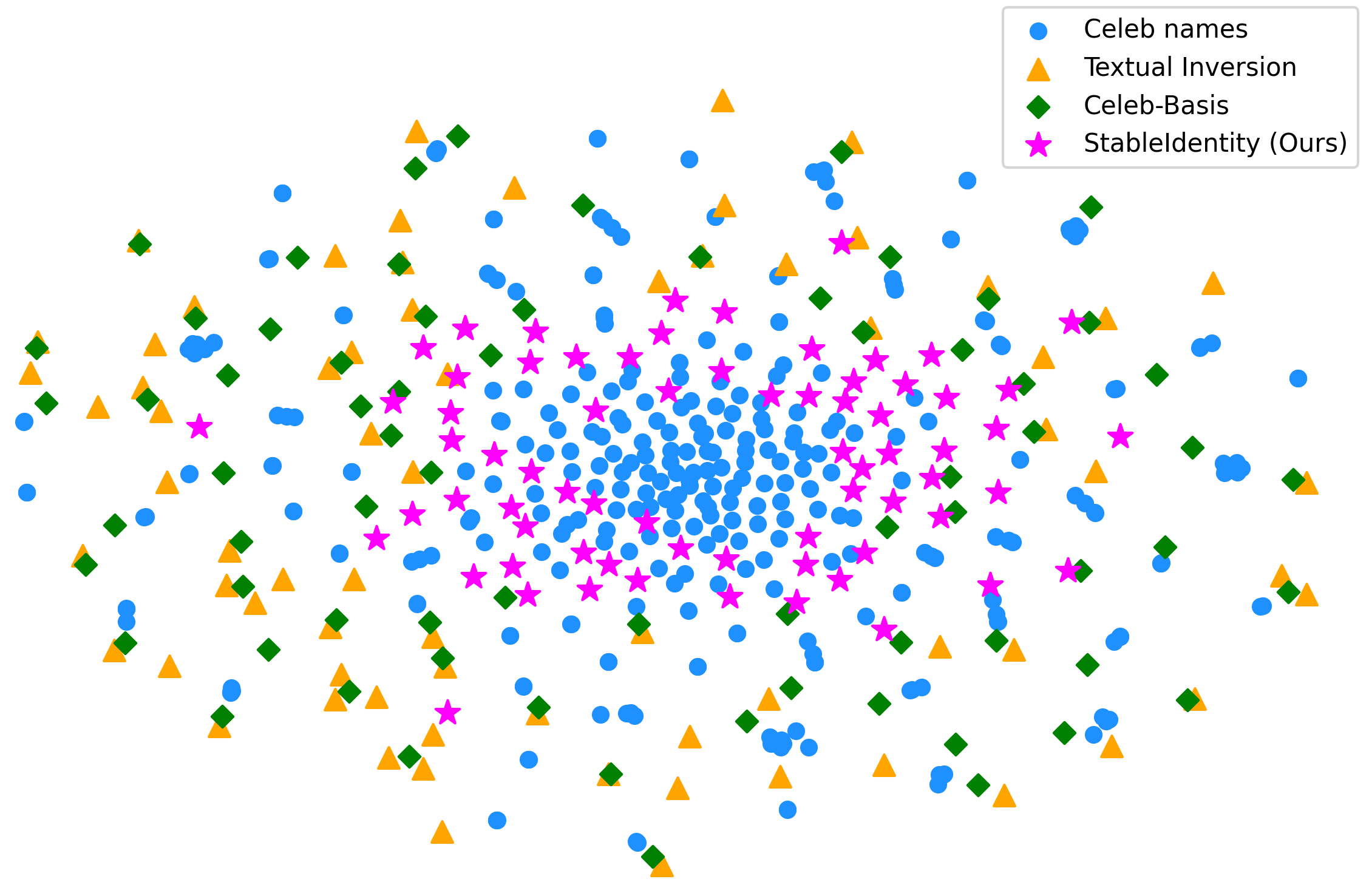"}
      \caption{2-D visualization of word embeddings using \textit{t}-SNE with Celeb names, Textual Inversion, Celeb-Basis and our method.}
  \label{fig:tsne}
\end{figure}

\subsection{Word-Embedding-Based Methods Analysis}
Considering that Textual Inversion, Celeb-Basis and our method are all optimized in the word embedding space, we further analyze 70 embeddings learned by these methods from different perspectives. To match the dimension of word embeddings, Textual Inversion is conducted with 2-word version and Celeb-Basis is implemented with SD2.1 for analysis.

\begin{table}[t]
\centering
\caption{Comparison with baselines optimized in the word embedding space on training time, maximum and minimum values of learned embeddings.}
\label{tab:time_max_min}
\setlength{\tabcolsep}{5pt}

\begin{tabular}{lccc}
\toprule
                           & Training time & Max    & Min     \\
\midrule                           
Celeb names                & $-$             & 0.0551 & -0.0558 \\
Textual Inversion          & 43mins         & 0.7606  & -0.9043 \\
Celeb-Basis                & 8mins          & 0.1592 & -0.1499 \\
StableIdentity~(Ours) & 4mins          & 0.0557 & -0.0520 \\
\bottomrule  
\end{tabular} 
\end{table}

\begin{figure}[!t]
  \centering
  \includegraphics[width=1\linewidth]{"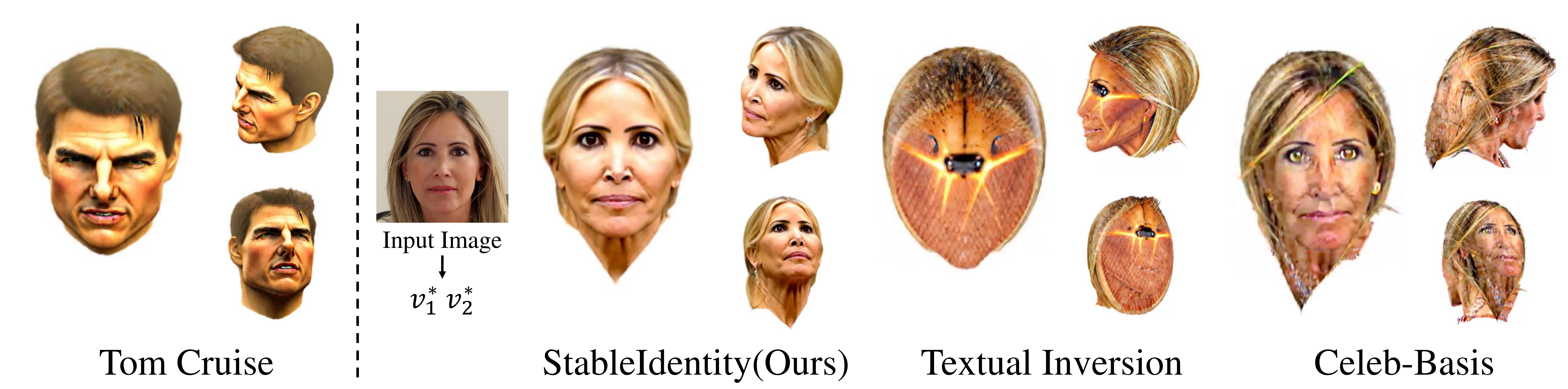"}
      \caption{Comparison of 3D generation based on LucidDreamer. We show the result of a celeb name ``Tom Cruise"~(prompt) as a standard and the results with the embeddings $[v^*_1, v^*_2]$ learned from competing methods~(\textit{Zoom-in for the best view}).}
  \label{fig:3d_compare}
\end{figure}

To intuitively show the difference between the distributions of learned embeddings and celeb embeddings, we use the \textit{t}-SNE~\cite{van2014accelerating} to visualize word embeddings in Figure~\ref{fig:tsne}. ``Celeb names" denotes the word embeddings corresponding to the collected 326 celeb names. It can be observed that the distribution of ours is more compact with fewer outliers and closer to the real distribution of celeb names, achieving the best identity-consistent editability. Besides, we compare the max \& min values of the learned embeddings and training time in Table~\ref{tab:time_max_min}. Our method is faster than existing methods of the same kind, and the value range is closest to real celeb embeddings.

Furthermore, to examine the generalization ability of these methods, we present 3D generation results with the learned identity embeddings directly using the mentioned 3D generation pipeline LucidDreamer in Figure~\ref{fig:3d_compare}. And we show a standard result using a celeb name ``Tom Cruise'' as a prompt. Obviously, our method achieves celeb-like results in every 3D view, which further demonstrates stable and strong generalization ability of our learned identity.

\section{Conclusion}
In this paper, we propose \textit{StableIdentity}, a customized generation framework which can inject anybody into anywhere. The model architecture that integrates identity and editability prior allows the learned identity to master identity-consistent recontextualization ability. Besides, the designed masked two-phase diffusion loss enables the learned identity more stable. Extensive quantitative and qualitative experiments demonstrate the superiority of the proposed method. Surprisingly, our method can directly work with the plug-and-play SD-based modules such as ControlNet, and even can insert the learned identity into off-the-shelf video/3D generated models without finetuning to produce outstanding effects. We hope that our work can contribute to the unification of customization over image, video, and 3D generation tasks.


\clearpage
\appendix

\twocolumn[{
\renewcommand\twocolumn[1][]{#1}
\begin{center}
\centering
    \captionsetup{type=figure}
    \includegraphics[width=1\linewidth]{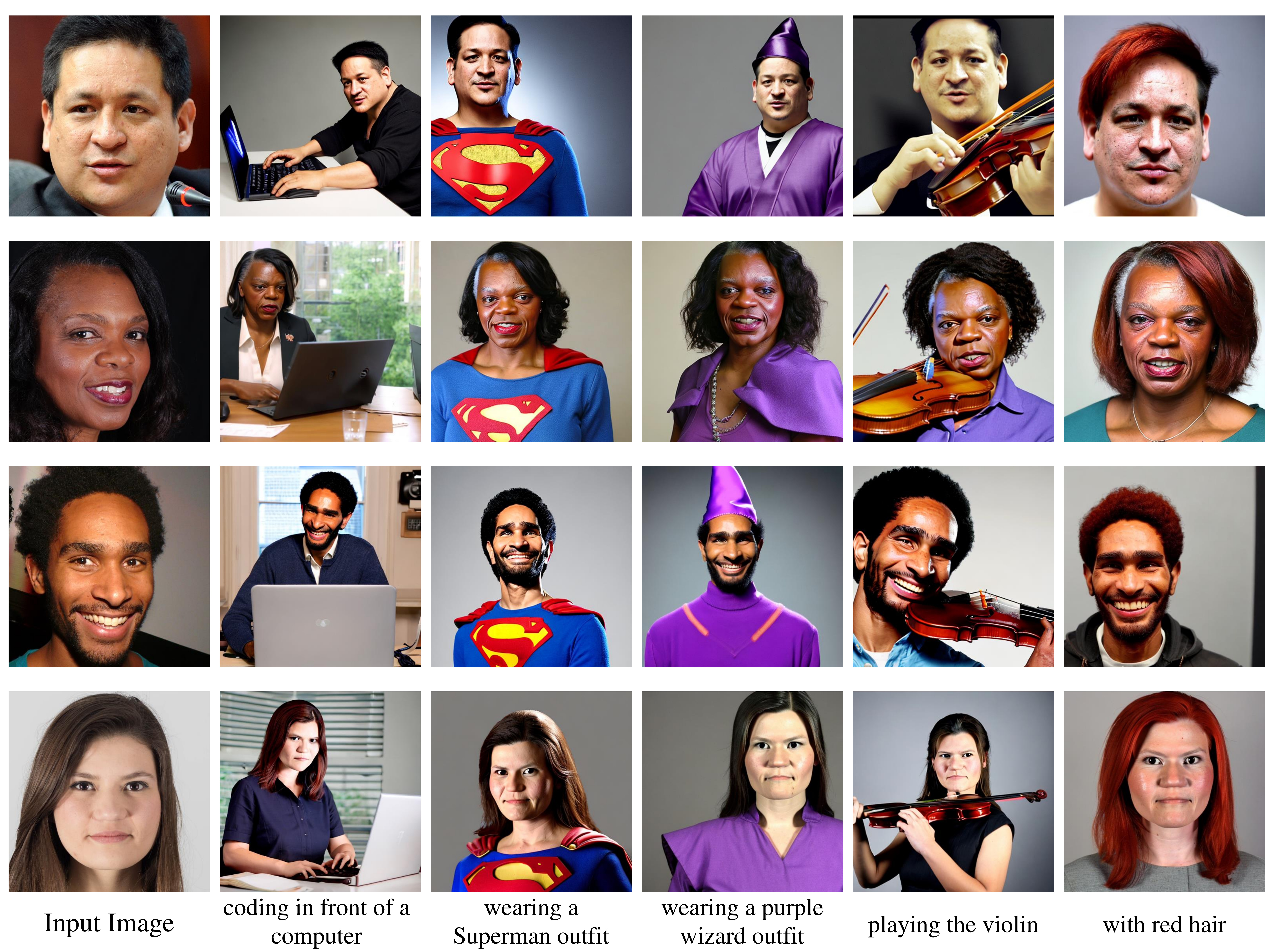}
    \captionof{figure}{More generated results with the proposed \textit{StableIdentity} for different identities~(including various races) under various contexts~(covering decoration, action, attribute).
    }
    \label{fig:supp_context}
    \vspace{0.6cm}
\end{center}
}]

\begin{figure*}[!t]
  \centering
  \includegraphics[width=1\linewidth]{"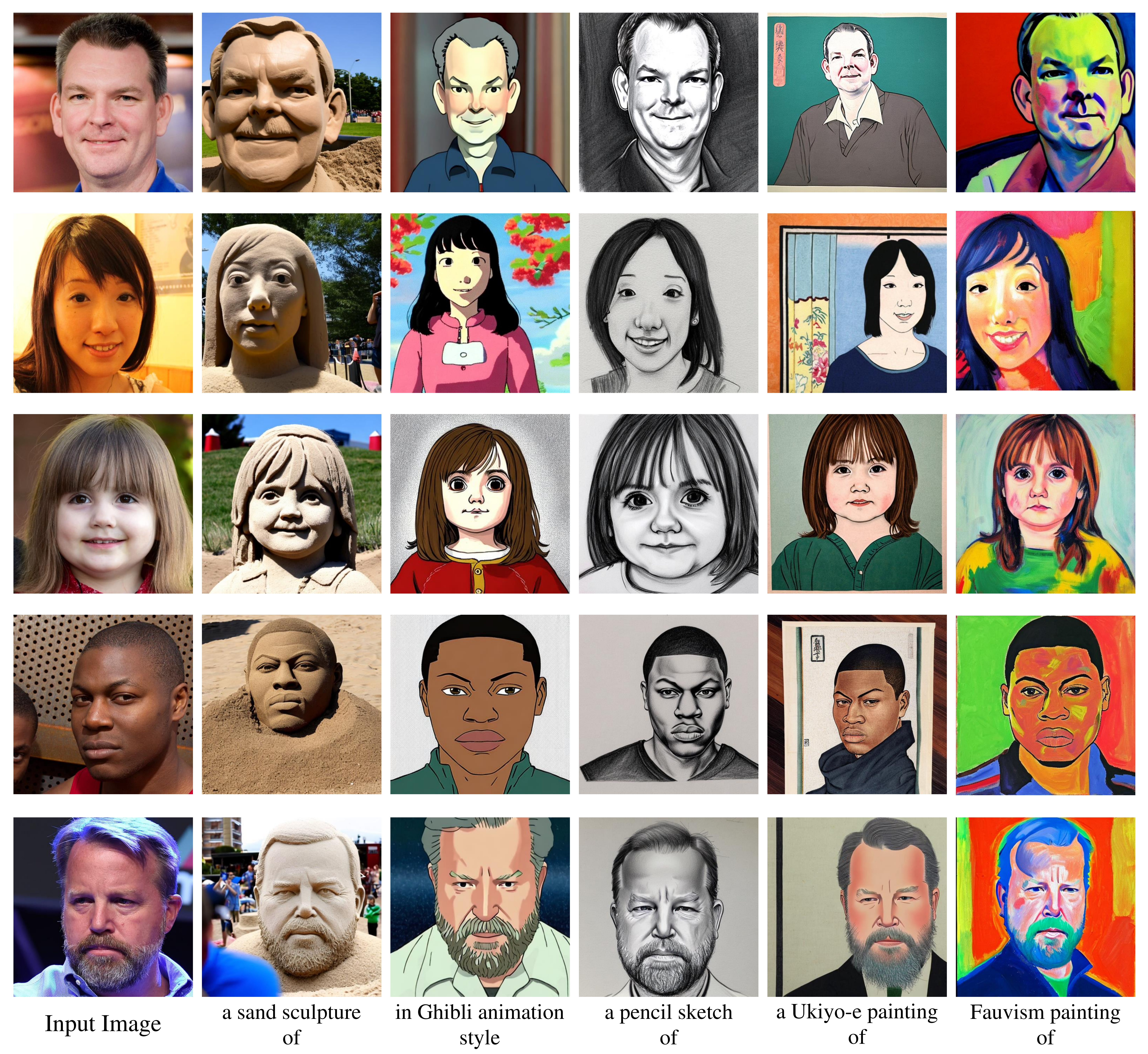"}
      \caption{Additional customized results with \textit{StableIdentity} for diverse artistic styles.}
  \label{fig:supp_style}
\end{figure*}

\section{More Visual Results}

\noindent\textbf{More Customized Results.}
As shown in Figure~\ref{fig:supp_context}, it can be observed that \textit{StableIdentity} can handle different races in various contexts. On the other hand, we also show the customized results with diverse artistic styles in Figure~\ref{fig:supp_style}. Overall, the generated results have satisfactory identity preservation and editability, which demonstrates the effectiveness of the proposed method.

\begin{figure*}[!t]
  \centering
  \includegraphics[width=1\linewidth]{"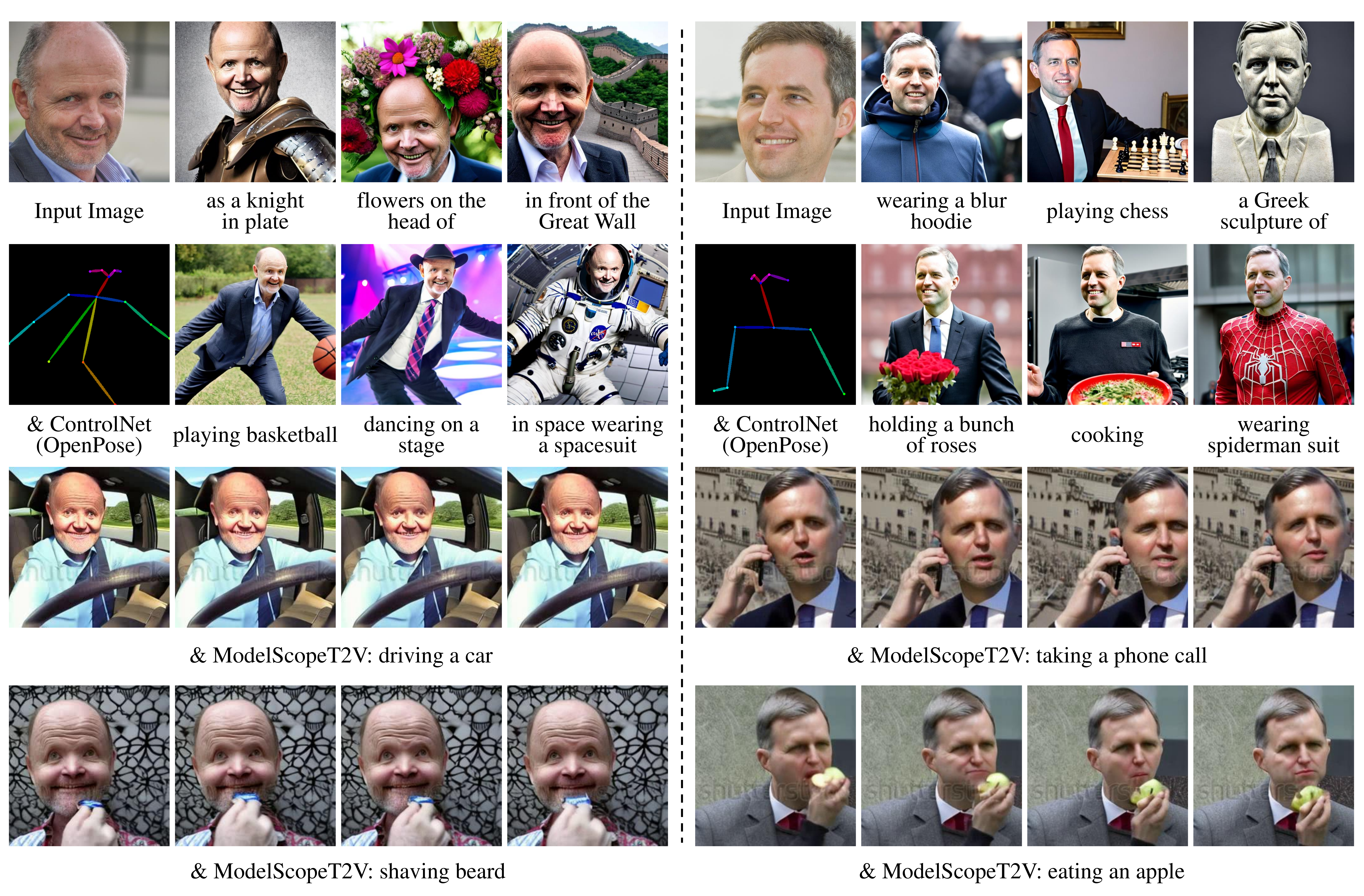"}
      \caption{Pose-controlled customized image generation~(StableIdentity~\&~ControlNet) and zero-shot identity-driven customized video generation~(StableIdentity~\&~ModelScopeT2V).}
  \label{fig:supp_controlnet_video}
\end{figure*}

\begin{figure*}[!t]
  \centering
  \includegraphics[width=1\linewidth]{"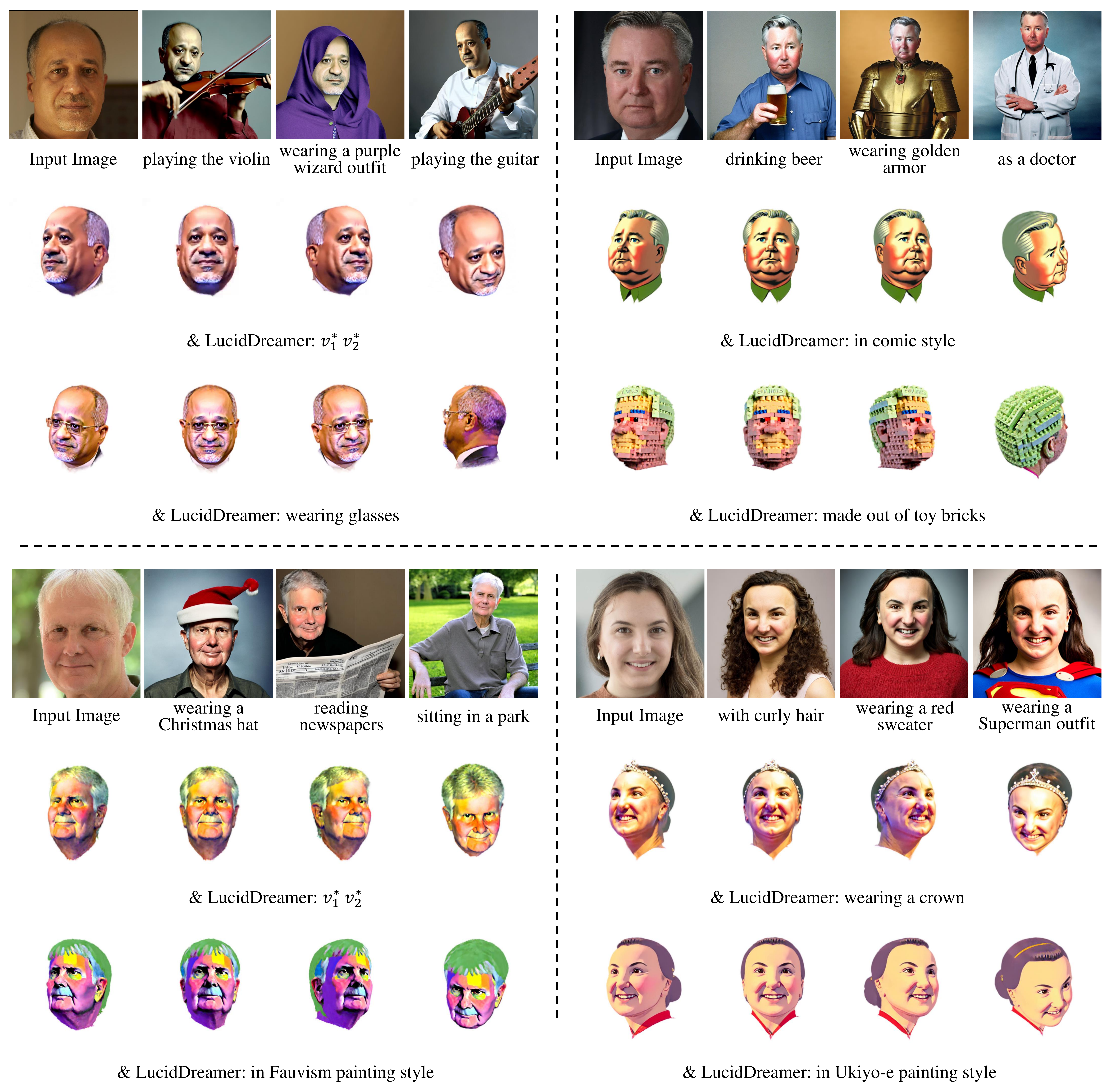"}
      \caption{Zero-shot identity-driven customized 3D generation~(StableIdentity~\&~LucidDreamer). As mentioned in Sec.~4.1, we omit the placeholders $v_1^*$ $v_2^*$ of prompts such as ``$v_1^*$ $v_2^*$ wearing glasses" for brevity. Here, we use ``$v_1^*$ $v_2^*$ as the input prompt to show the 3D reconstruction for the learned identities.}
  \label{fig:supp_3d}
\end{figure*}

\begin{figure*}[!t]
  \centering
  \includegraphics[width=1\linewidth]{"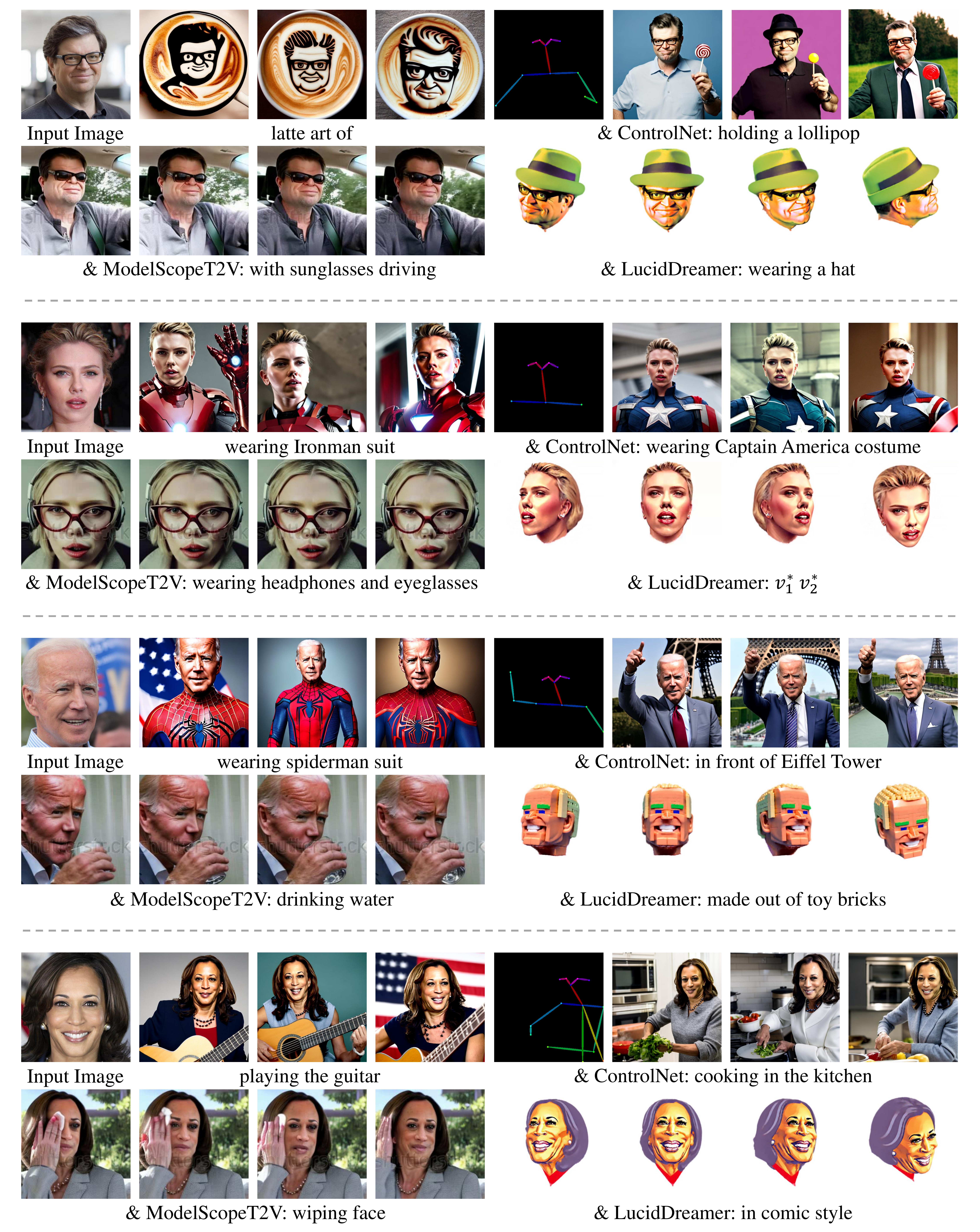"}
      \caption{More image/video/3D customized generation results for celeb photos as input.}
  \label{fig:celeb_photo}
\end{figure*}

\noindent\textbf{StableIdentity \& Image/Video/3D Models.} 
In addition, as shown in Figure~\ref{fig:supp_controlnet_video},~\ref{fig:supp_3d}, we show more image/video/3D customized generation results with ControlNet~\cite{zhang2023adding}, ModelScopeT2V~\cite{wang2023modelscope}, and LucidDreamer~\cite{liang2023luciddreamer}. As introduced in Sec.~\ref{sec:implement_details}, StableIdentity can be considered as introducing new identity for the dictionary of CLIP text encoder. Therefore, the learned identity can be naturally inserted into various contexts or even into video/3D generated models for identity-driven customized generation. Due to the limited performance of 3D generation, following~\cite{liang2023luciddreamer}, we only generate and edit in the head region, which can clearly demonstrates whether the learned identity is successfully inserted or not. Impressive experimental results show that our method can be stably injected into image/video/3D generative models to achieve identity-consistent recontextualization. Furthermore, we also show more customized generation results using celeb photos as input as shown in~\ref{fig:celeb_photo}.

\begin{figure}[!t]
  \centering
  \includegraphics[width=1\linewidth]{"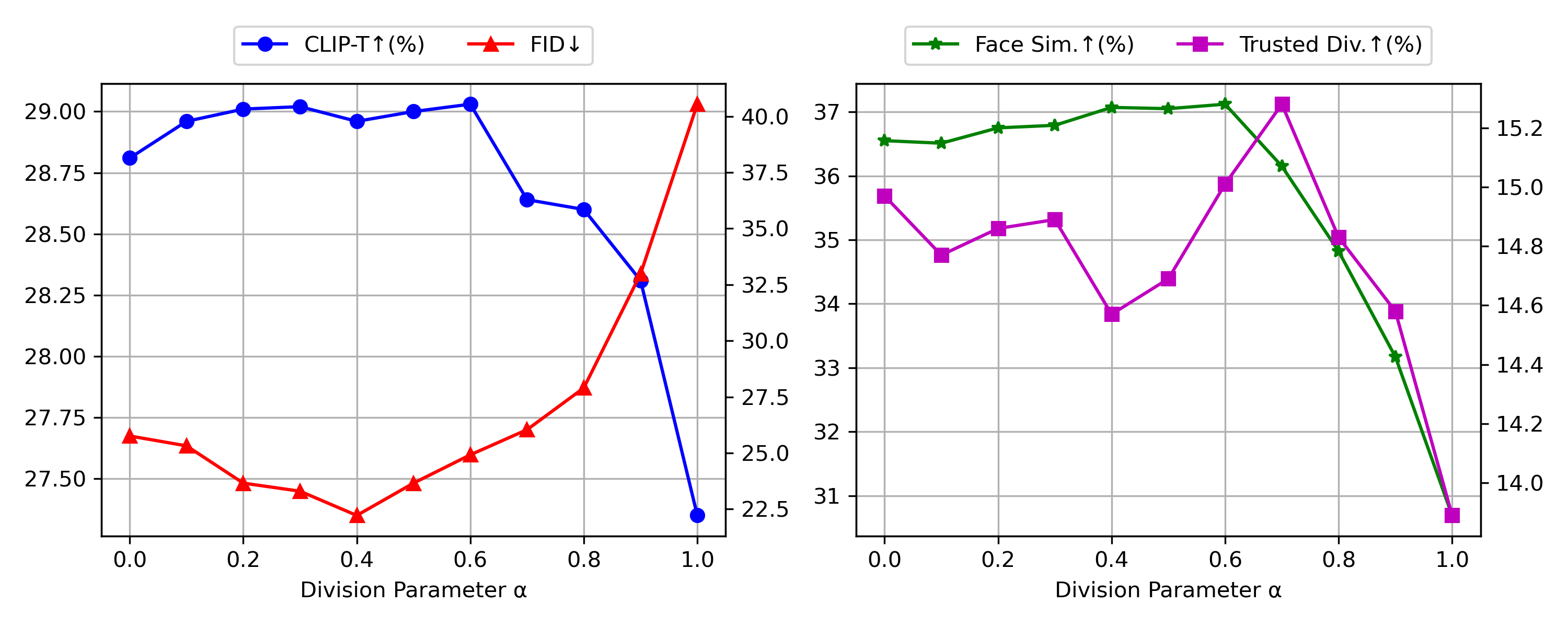"}
      \caption{Parameter analysis for the division parameter $\alpha$.}
  \label{fig:alpha}
\end{figure}

\begin{figure*}[!t]
  \centering
  \includegraphics[width=1\linewidth]{"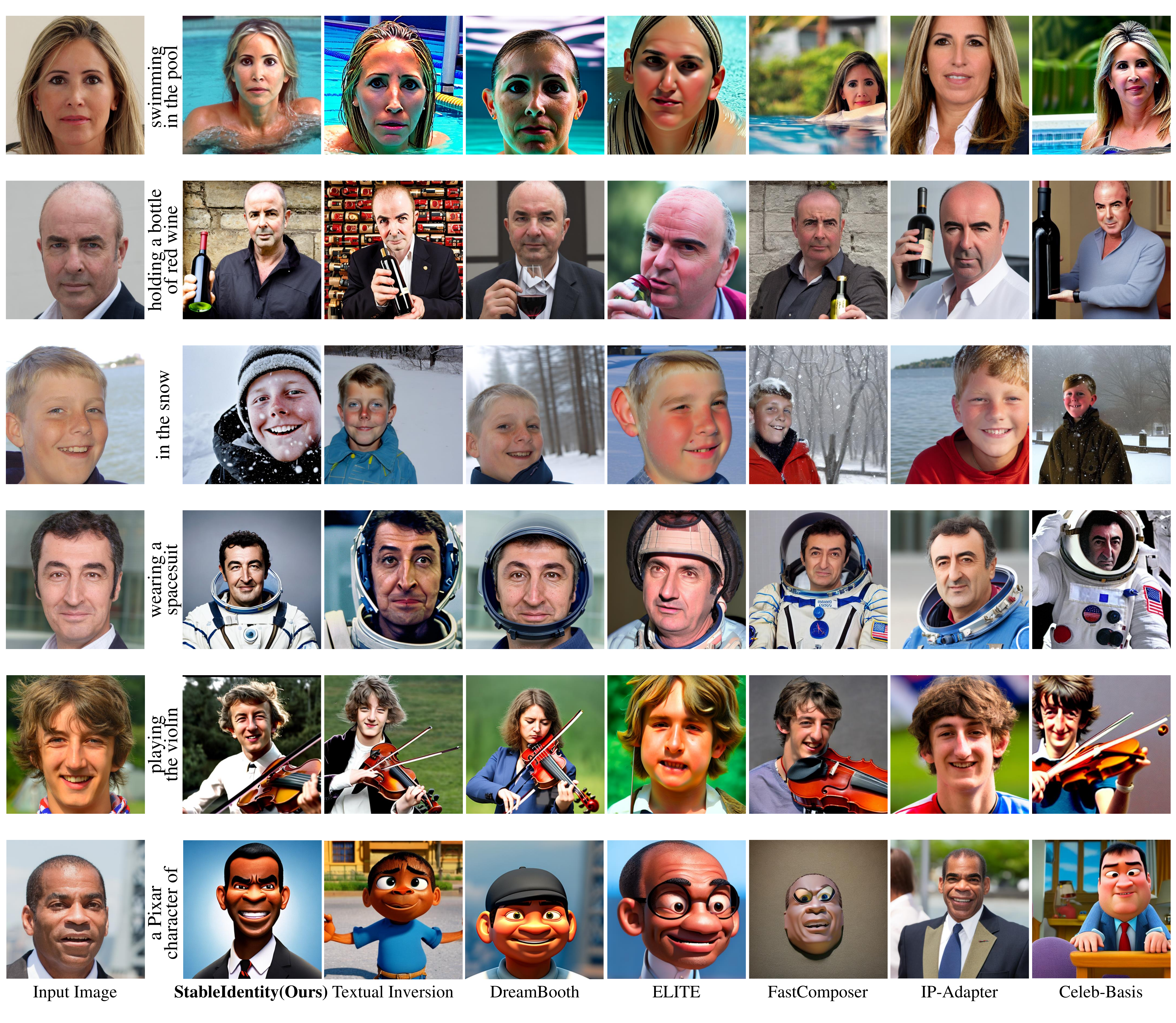"}
      \caption{More qualitative comparisons for different identities~(including various races) with diverse text prompts~(covering decoration, action, background, style). Our method shows best performance for identity preservation and editability~(\textit{Zoom-in for the best view}).}
  \label{fig:supp_comparison}
\end{figure*}

\noindent\textbf{More Comparisons.}
As shown in Figure~\ref{fig:supp_comparison}, we further compare with baselines on decoration, action, background, style. Obviously, we achieve the best results for identity preservation, editability and image quality. DreamBooth, which seems to perform well~(row 1,3,4), either overfits to the input image or fails to produce results similar to the target identity.

\section{Implementation Details}
\noindent\textbf{Filtering Celeb Names.} As mentioned in Sec.~\ref{related_work}, Celeb-Basis~\cite{yuan2023inserting} collects 691 celeb names which are editable in Stable Diffusion~\cite{rombach2022high}. We only filter out names consisting of only two words and then count the number of the corresponding tokens. The percentages of 2 words$\rightarrow$\{2,3,4,5,6\} tokens are 56\%, 27\%, 13\%, 3\%, and 0.3\% respectively. To obtain a more accurate prior space, we choose the setting of 2 words$\rightarrow$2 tokens, which has more sampling points.

\noindent\textbf{Division Parameter $\alpha$.}
As shown in Figure~\ref{fig:alpha}, we present the effect of different $\alpha$ in $[0,0.1,\cdot\cdot\cdot, 1]$. Empirically, $\alpha\in[0.4,0.6]$ shows better identity preservation, editability and image quality. When $\alpha$ is larger, meaningless reconstruction will lead to performance degradation.

\section{Limitations}
Although the proposed method achieves outstanding performance for customization generation of new identities and can collaborate with the off-the-shelf image/video/3D models, it still faces some limitations. (1)~Since we only act in the word embedding space and fix the Stable Diffusion~(SD), we inherit not only the excellent performance of SD, but also some drawbacks, such as hand anomalies~\cite{zhang2023detecting}.  (2)~Existing text-to-video generation models can generate with diverse contexts, but is still immature for human-centric generation~\cite{wang2023modelscope}. It leads to limited performance for the video customization generation.

\end{document}